\documentclass[sigconf,screen]{acmart}
\usepackage{amsthm}
\usepackage{hyperref}
\usepackage{multirow}
\usepackage{amsmath}
\usepackage{booktabs}
\usepackage[skip=0.5\baselineskip,font=bf]{caption}
\usepackage{courier}
\usepackage{graphicx} 
\usepackage[export]{adjustbox}
\usepackage{subcaption}
\usepackage{eqnarray}
\usepackage{bm}
\usepackage{bbm}
\usepackage{amsmath}
\usepackage{amsthm}
\usepackage{mathtools}
\usepackage{multirow}
\usepackage{xcolor}
\usepackage{colortbl}
\usepackage{wrapfig}
\usepackage{xcolor}
\usepackage{dsfont}
\usepackage{amsthm}
\theoremstyle{definition}

\usepackage[capitalise]{cleveref}

\newcommand{\cell}[2]{\setlength{\tabcolsep}{0pt}\begin{tabular}{#1}#2 \end{tabular}}

\newcommand{\textds}[1]{{\footnotesize{\texttt{#1}}}}

\theoremstyle{plain}
\newtheorem{theorem}{Theorem}
\newtheorem{lemma}{Lemma}

\newcommand\Set[1]{\{#1\}}

\newcommand{\Probability}{\mathbb{P}}
\newcommand{\Indicator}{\mathds{1}}
\newcommand{\Hypotheses}{\mathcal{H}}
\newcommand{\Expectation}{\mathbb{E}}
\newcommand{\Reals}{\mathbb{R}}
\newcommand{\SuchThat}{\textnormal{s.t.}}
\newcommand{\WithProb}{\textrm{w.p.}}

\DeclareMathOperator{\TPR}{TPR}
\DeclareMathOperator{\FPR}{FPR}

\newcommand{\noisy}{\tilde}
\newcommand{\noisyeps}{\hat{\epsilon}}
\DeclareMathOperator*{\argmin}{argmin}
\DeclareMathOperator{\sign}{sign}

\newcommand\numberthis{\addtocounter{equation}{1}\tag{\theequation}}
\newcommand{\squishlist}{
\begin{list}{{{\small{$\bullet$}}}}
{\setlength{\itemsep}{3pt}      \setlength{\parsep}{1pt}
\setlength{\topsep}{1pt}       \setlength{\partopsep}{0pt}
\setlength{\leftmargin}{1em} \setlength{\labelwidth}{1em}
\setlength{\labelsep}{0.5em} } }
\newcommand{\squishend}{  \end{list}  }

\usepackage{scalerel,stackengine}
\stackMath
\newcommand\reallywidehat[1]{%
\savestack{\tmpbox}{\stretchto{%
  \scaleto{%
    \scalerel*[\widthof{\ensuremath{#1}}]{\kern-.6pt\bigwedge\kern-.6pt}%
    {\rule[-\textheight/2]{1ex}{\textheight}}
  }{\textheight}%
}{0.5ex}}%
\stackon[1pt]{#1}{\tmpbox}%
}

\copyrightyear{2021}
\acmYear{2021}
\setcopyright{rightsretained}
\acmConference[FAccT '21]{Conference on Fairness, Accountability, and
Transparency}{March 3--10, 2021}{Virtual Event, Canada}
\acmBooktitle{Conference on Fairness, Accountability, and Transparency
(FAccT '21), March 3--10, 2021, Virtual Event,
Canada}\acmDOI{10.1145/3442188.3445915}
\acmISBN{978-1-4503-8309-7/21/03}



\begin{document}

\title{Fair Classification with Group-Dependent Label Noise}

\author{Jialu Wang}
\email{faldict@ucsc.edu}
\affiliation{
    \institution{UC Santa Cruz}
    \city{Santa Cruz}
    \state{CA}
    \country{USA}
}

\author{Yang Liu}
\email{yangliu@ucsc.edu}
\affiliation{
    \institution{UC Santa Cruz}
    \city{Santa Cruz}
    \state{CA}
    \country{USA}
}

\author{Caleb Levy}
\email{cclevy@ucsc.edu}
\affiliation{
    \institution{UC Santa Cruz}
    \city{Santa Cruz}
    \state{CA}
    \country{USA}
}

\begin{abstract}
This work examines how to train fair classifiers in settings where training labels are corrupted with random noise, and where the error rates of corruption depend both on the label class and on the membership function for a protected subgroup. Heterogeneous label noise models systematic biases towards particular groups when generating annotations. We begin by presenting analytical results which show that naively imposing parity constraints on demographic disparity measures, without accounting for heterogeneous and group-dependent error rates, can decrease both the accuracy \emph{and} the fairness of the resulting classifier. Our experiments demonstrate these issues arise in practice as well. We address these problems by performing empirical risk minimization with carefully defined surrogate loss functions and surrogate constraints that help avoid the pitfalls introduced by heterogeneous label noise. We provide both theoretical and empirical justifications for the efficacy of our methods. We view our results as an important example of how imposing fairness on biased data sets without proper care can do at least as much harm as it does good.
\end{abstract}

\maketitle

\section{Introduction}\label{Sec::Introduction}

Recent work shows that machine learning classifiers can perpetuate and amplify existing systemic injustices in society. Notable examples include discrepancies in allocation of medical care to patients on the basis of race~\cite{Obermeyer_2019} and significant disparities in predicting recidivism rates for African-American defendants~\cite{angwin2016machinebias,Chouldechova_2016}, and more \cite{vayena2018machine,popejoy2016genomics,buolamwini2018gender}. A number of techniques have been developed in order to mitigate bias in machine learning classifiers~\cite{zafar2015fairness,feldman2015certifying,hardt2016equality,agarwal2018reductions,menon2018cost,celis2019classification}. Typically, these methods consider populations with groups corresponding to a set of protected sensitive attributes, such as race or gender. The classifier is then required to exhibit similar behavior across all groups~\cite{zafar2015fairness,hardt2016equality,Chouldechova_2016,kamiran2009classifying}. This can be done by imposing equality of true positive rate or true negative rate conditioned on group membership. These are called ``fairness,'' or parity constraints.

Many of these methods assume the availability of clean and accurate labels. However, this is often not the case. In fact, bias in data is particularly pertinent to label corruption. To make things worse, the accuracy of available labels is often strongly influenced by whether a person falls within a protected group, and these discrepancies can have significant and often life-altering outcomes. For example, it has been shown that labels for criminal activity generated via crowdsourcing are systematically biased against certain racial groups~\cite{dressel2018accuracy}. As another example, both women and lower-income individuals often receive significantly less accurate diagnoses for cancer and other ailments than men, due to imbalance in the sample population of medical trials~\cite{Gianfrancesco_2018}, and due to bias from doctor treatment~\cite{Tumor_2016}. Similar discrepancies arise in the accuracy of mathematical aptitude evaluations for males and females in primary school~\cite{Lindberg_2010}, and it has long been known that an employer's evaluation of a resume will be influenced by the perceived ethnic origin of an applicant's name~\cite{Bertrand_2004}. Moreover, studies show that people of all races use and sell illegal drugs at remarkably similar rates, but in some states, black male have been admitted to prison on drug charges at rates twenty to fifty times greater than those of white men~\cite{Alexander2010TheNJ}.

The structure and magnitude of group-specific label noise can dramatically affect the performance \emph{and} fairness of a classifier. To see this, we consider the following examples.
\paragraph{Example~\ref{table:NoiseReducesAccuracy}.} \emph{Enforcing fairness constraints without accounting for group-specific label noise can harm the accuracy of the classifier for the group whose labels have been accurate recorded.}
  
Consider training classifiers using data from two groups $z\in\Set{A,B}$ with homogeneous data distributions $\Probability(Y =+1|X=\bm{x},z=A) = \Probability(Y=+1|X=\bm{x},z=B)$, where $\bm{x}=[x_1,x_2]$, a 2-dimensional feature vector. In this setting, the Bayes-optimal classifiers for $A$ and $B$ (denoted as $f^*_A$ and $f^*_B$ respectively) will obey any parity constraint. However, suppose group $A$ has a set of clean labels, while group $B$ has clean labels when the ground truth is $y=+1$ but there is a 70\% chance that corrupting noise will cause the observed label to be flipped from the true value when $y=-1$. In this case, $f^*_{fair}$ trained on both groups achieves perceived equal True Positive Rates (TPR) (50\%) between the two groups and is the best one to do so - this indeed hurts group $A$'s prediction performance (as opposed to 100\% accuracy before), but the labels in group $A$ are not affected by noise. Although \cite{Blum2019RecoveringFB} also considers this single-group noise setting and shows that fairness interventions could aid in reducing the error caused by label bias, our observation demonstrates a special case where potential harm occurs.
\captionsetup[table]{name=Example}
\begin{table}[ht]
\centering
\caption{ 
A simple classification problem to illustrate the possibility of harming the clean group when training a fair True Positive Rate (TPR) model over a set of noisy labels. }
\label{table:NoiseReducesAccuracy}
\resizebox{\linewidth}{!}{
\begin{tabular}{c*{3}{>{\hspace{0.5em}}rrr}}
& \multicolumn{3}{c}{\textsc{Group $A$}}
& \multicolumn{3}{c}{\textsc{Group $B$}}
& \multicolumn{3}{c}{\textsc{Pooled}} \\ 
\cmidrule(lr){2-4} 
\cmidrule(lr){5-7} 
\cmidrule(lr){8-10}

$(x_1, x_2)$,~~~$y$ &  
$+1$ & $-1$ & ${f^*_A}$ & 
$+1$ & $-1$ & ${f^*_B}$ &
$+1$ & $-1$ & ${f^*_{fair}}$ \\ 

\cmidrule(lr){1-1} 
\cmidrule(lr){2-4} 
\cmidrule(lr){5-7} 
\cmidrule(lr){8-10}

$(0, 0),~~-1$ & 
0 & 25 & $-1$ &
70 & 30 & $+1$ &
70 & 55 & $+1$ \\ 

$(0, 1),~~-1$ & 
0 & 25 & $-1$ &
70 & 30 & $+1$ &
70 & 55 & $-1$ \\ 

$(1, 0),~~+1$ & 
25 & 0 & $+1$ &
100 & 0 & $+1$ &
125 & 0 & $+1$ \\ 

$(1, 1),~~+1$ & 
25 & 0 & $+1$ &
100 & 0 & $+1$ &
125 & 0 & $-1$ \\ 

\cmidrule(lr){1-1} 
\cmidrule(lr){2-4} 
\cmidrule(lr){5-7} 
\cmidrule(lr){8-10}
\end{tabular}
}
\end{table}

\begin{table}[ht]
\centering
\caption{A simple classification problem to illustrate the possibility of
wrongly perceived fairness due to training on noisy labels.}
\resizebox{\linewidth}{!}{
\begin{tabular}{c*{3}{>{\hspace{0.5em}}rrr}}
& \multicolumn{3}{c}{\textsc{Group $A$}}
& \multicolumn{3}{c}{\textsc{Group $B$}}
& \multicolumn{3}{c}{\textsc{Pooled}} \\ 
\cmidrule(lr){2-4} 
\cmidrule(lr){5-7} 
\cmidrule(lr){8-10}

$(x_1, x_2)$,~~$(y_A,y_B)$ &  
$+1$ & $-1$ & ${f^*_A}$ & 
$+1$ & $-1$ & ${f^*_B}$ &
$+1$ & $-1$ & ${f^*_{fair}}$ \\ 

\cmidrule(lr){1-1} 
\cmidrule(lr){2-4} 
\cmidrule(lr){5-7} 
\cmidrule(lr){8-10}

$(0, 0),~~(-1,-1)$ & 
0 & 100 & $-1$ &
75 & 225 & $-1$ &
75 & 325 & $-1$ \\ 

$(0, 1),~~(-1,+1)$ & 
0 & 100 & $-1$ &
75 & 25 & $+1$ &
75 & 125 & $+1$ \\ 

$(1, 0),~~(+1,+1)$ & 
100 & 0 & $+1$ &
75 & 25 & $+1$ &
175 & 25 & $-1$ \\ 

$(1, 1),~~(+1,+1)$ & 
100 & 0 & $+1$ &
75 & 25 & $+1$ &
175 & 25 & $+1$ \\

\cmidrule(lr){1-1} 
\cmidrule(lr){2-4} 
\cmidrule(lr){5-7} 
\cmidrule(lr){8-10}
\end{tabular}
}

\label{table:NoiseReducesFairness}
\end{table}
\captionsetup[table]{name=Table}
\paragraph{Example~\ref{table:NoiseReducesFairness}.} \emph{A classifier may appear to achieve parity when it does not. Furthermore, imposing a parity constraint might actually make everyone worse off.}

Consider training classifiers using data from two groups $z\in\Set{A,B}$ with heterogeneous data distributions $\Probability(Y=+1\mid X=\bm{x},z=A) = \Probability(Y=+1\mid X=\bm{x},z=B)$. Suppose group $A$ has a set of clean labels, while one quarter of group $B$'s labels are incorrect. We denote the Bayes-optimal classifiers for $A$ and $B$ as $f^*_A$ and $f^*_B$ respectively and they obey any parity constraint. The classifier $f^*_{fair}$ trained on the observed corrupted data is subject to equal TPR constraint for both groups. \footnote{Note that $f^\ast_{fair}$ on the pooled data output +1 for $(0,1)$ and -1 for $(1,0)$ because equal TPR constraint is enforced. In this case, the TPRs for both groups are 50\%. If the classifier output -1 for $(0,1)$ and +1 for $(1,0)$ instead, the TPR for group A is 100\% while the TPR for group B is only 50\%, which violates the equal TPR constraint.} However, $f^*_{fair}$ has a higher TPR ($2/3$: $200$ correct predictions out of $300$ true +1 labels) on $B$ than on $A$ ($1/2$: $100$ correct predictions out of $200$ true +1 labels) when evaluated on the clean data.

In this paper, we look at the problem of fair classification from data whose labels are corrupted, such that the error rates of corruption are group-dependent. Several recent works deal with fair classification with noisy labels~\cite{jiang2019identifying,Lamy2019NoisetolerantFC,Blum2019RecoveringFB}. In particular, it has been shown that fairness constraints on the noisy training labels can be beneficial when the label noise is homogeneous across the different groups that are to be protected~\cite{Blum2019RecoveringFB}. More recently, \cite{pmlr-v108-fogliato20a} shows that how the true fairness rates, such as TPR, are related to observed quantities with respect to noise parameters. Our work complements these results: we show that enforcing fairness constraints when training on data with noisy labels produces a classifier that violates the fairness constraints as measured with respect to the clean data.
We then provide a fair empirical risk minimization (ERM) framework that handles heterogenous label noise. Our framework uses an estimation procedure that infers the knowledge of group-dependent noise in the training data and applies this knowledge using bias removal techniques, thus eliminating the effects of noisy labels in both the objective function and the fairness constraints (in expectation).

Our main contributions are as follows: 
(1) We show that imposing fairness constraints on the training process without accounting for bias in the noisy labels can result in classifiers being less accurate \emph{and} less fair (Theorems~\ref{thm:NoiseHarmsAccuracy} and~\ref{thm:NoiseHarmsFairness} of Section~\ref{sec:Analysis}). (2) We experimentally demonstrate that these harms can indeed occur in practice with real data sets, and show that obliviously enforcing equality of opportunity without awareness of the noise leads to classifiers with no discriminatory power. (3) We design two noise-resistant fair ERM approaches that address these problems (Section~\ref{sec:fairERM}). The main idea is to construct unbiased estimators of the loss functions and of the fairness constraints. (4) We provide empirical evidence showing that these fair ERM solutions improve both accuracy and fairness guarantees when facing group-dependent label noise (Section~\ref{Sec::Experiments}). (5) Our codes for solving the noise-resistant fairness constrained ERM can be found at \url{https://github.com/Faldict/fair-classification-with-noisy-labels}.

\subsection{Related Works}
A great deal of research has been devoted to fair classification in general, including fair classification under statistical constraints \cite{zafar2015fairness,feldman2015certifying,hardt2016equality,agarwal2018reductions}, decoupled training with preference guarantees \cite{zafar2017fairness,dwork2017decoupled,lipton2018disparate,Ustun_2019,celis2019classification}, and preventing gerrymandering \cite{kearns2017preventing}, among many others \cite{menon2018cost,chen2018my}. 

In this work, we specifically focus on fairness in the presence of biased and group-dependent noisy training labels. Our work contributes to the fair classification literature by introducing robust methods for dealing with heterogeneous label noise. We also provide insight into the effects of noise being present in the labels. Our work parallels others' on fair classification with noisy labels~\cite{jiang2019identifying,Blum2019RecoveringFB}. Ours differs primarily in two main respects. First, existing works often assume knowledge of the noise generation process. Second, previous works have only considered noise rates that are homogeneous across different groups. We consider a more realistic setting, where different groups might suffer different levels of bias, and therefore reach very different conclusions. Mitigating bias is substantially more challenging in our setting. Nevertheless, our results could generalized prior work when the noise is assumed constant across groups, or only one group is assumed to have noise.

Both of our fair ERM approaches extend the literature on learning with noisy data~\citep{angluin1988learning,Manwani_2013,natarajan2013learning,Frenay_2014,scott2015rate,menon2015learning,liu2016classification,patrini2017making,charoenphakdee2019symmetric}. Our first uses surrogate loss functions based on~\cite{natarajan2013learning} to create unbiased estimators of the fairness constraints. This first approach requires knowledge of the noise parameters. Our second approach relaxes this assumption by extending the work of~\cite{Liu_2019} to account for both biases in the fairness constraints and for group specific label noise.

Recent work on fair classification with imperfect data shows how to emulate noiseless fair classification by appropriately re-scaling the fairness tolerance with the noise but is only restricted to class-conditional random noise without considering group difference~\cite{Lamy2019NoisetolerantFC}. Most of the reported results are for the cases with noisy sensitive attributes but not the labels (despite that the authors provided discussions to how the two problems are related). The surrogate fairness constraints in our paper could be viewed as an extension of their method. Nonetheless, our work is more general, as we consider the more sophisticated settings with group-dependent label noise. \cite{gupta2018proxy} explores the use of proxy variables when the sensitive attributes are missing. Lastly, \cite{pmlr-v108-fogliato20a} also provides some insights on correcting for observed predictive bias might further increase outcome disparities but is concerned with fairness evaluation rather than learning. In contrast with their work, we simplify the assumption on instance-dependent noise into group-dependent, and further develop two fair ERM approaches in terms of the unbiased estimators.
\section{Preliminaries}\label{sec:Preliminaries}
We start with a dataset with $n$ examples $(\bm{x}_i,y_i,z_i)_{i=1}^n$, where each example consists of a \emph{feature vector} $\bm{x}_i = (1,x_{i,1},\ldots,x_{i,d}) \in \Reals^{d+1}$, a \emph{label} $y_i \in \Set{+1,-1}$, and a \emph{group attribute} $z_i \in Z$ (e.g., $z_i=\Indicator[\texttt{\small{female}}]$). We assume that there are $m = |Z| \ge 2$ groups. We let $n_z$ denote the number of examples in group $z$, and we use $I_z= \Set{i \mid z_i=z}$ and $I = \bigcup_{z\in Z} I_z$ to denote their indices. We assume that each example is drawn iid from a joint distribution $\mathcal{D}$ of random variables $(X,Y,Z)$.

We use the data set to train a classifier $f\in\Hypotheses:\Reals^{d+1} \rightarrow \Set{+1,-1}$, where $\Hypotheses$ denotes our concept class. To this end, we consider solving a standard risk minimization problem with fairness constraints.
\begin{align}
\min_{f\in\Hypotheses} & ~~\Expectation_{(X,Y)\sim\mathcal{D}} \left[\Indicator(f(X) \ne Y)\right] ~~~~\\
\SuchThat & ~~~~ |F_z(f)-F_{z'}(f)|\le \delta \qquad \forall z,z'\in Z. \label{Con::FairnessConstraint}
\end{align}

Here, $F_z(f)$ is some fairness statistic of $f$ for group $z$ given the true labels $y$, such as \emph{true positive rate} :
$$ \text{(TPR)}: F_z(f) = \Probability(f(X)=+1|Y=+1,Z=z). 
$$
Constraint~\eqref{Con::FairnessConstraint} restricts the disparity between $z, z'$ to at most $\delta \ge 0$. A standard approach for performing above constrained minimization is via empirical risk minimization (ERM):
\begin{align}
\min_{f\in\Hypotheses} & ~\sum_{i=1}^n \Indicator(f(\bm{x}_i) \ne y_i)~~~~\label{eqn:risk:emp}\\ 
\SuchThat &~~~~ |\widehat{F}_z(f)-\widehat{F}_{z'}(f)|\le \delta \qquad \forall z,z'\in Z. \label{Con::FairnessConstraint:emp}
\end{align}
where $\widehat{F}_z(f)$ is our fairness metric defined using training data. For instance, when using the TPR as a fairness measure:
$$
  \widehat{F}_z(f) := 
        \frac
            {\#(f(\bm{x}_i)=+1, y_i=+1, z_i=z)}
            {\#(y_i=+1, z_i=z)},
$$
where $\#(\cdot)$ is simply a counting function that counts the number of samples that satisfy the specified conditions.

For computational purposes, ERM is performed in practice by minimizing over a classification-calibrated loss function \cite{bartlett2006convexity} $\ell:\Reals\times\Set{\pm 1}\rightarrow\Reals_+.$  This fits:
\begin{align}
\min_{f\in\Hypotheses} & ~\sum_{i=1}^n \ell(f(\bm{x}_i), y_i)~~~~\label{eqn:risk:surrogate}\\ 
\SuchThat &~~~~ |\widehat{F}_z(f)-\widehat{F}_{z'}(f)|\le \delta \qquad \forall z,z'\in Z. \label{Con::FairnessConstraint:surrogate}
\end{align}
Typical $\ell(\cdot)$s include square loss, logistic loss, cross-entropy loss and more.

We aim to train a classifier using a dataset where the ground truth labels $y_i$ are replaced by \emph{noisy} (or \emph{corrupted}) labels $\noisy{y}_i\sim \noisy{Y}$. A noisy label $\noisy{y}$ corresponds to a true label $y$ that may have been flipped based on noise rate $0\le \epsilon^+_z + \epsilon^-_z < 1$ (as a function of true label $y$). More precisely, we assume that the noise rates vary based on the true label $y$ as well as the group attribute $z$:
\begin{align*}
    \epsilon^+_z &= \Probability(\noisy{Y}=-1 \mid Y=+1, Z=z),~ \\
    \epsilon^-_z &= \Probability(\noisy{Y}=+1 \mid Y=-1, Z=z)
\end{align*}
i.e., the training labels are generated as:
\begin{align*}
\noisy{y}_i =
    \begin{cases}
        y_i               & \WithProb~ 1 - \epsilon^{\sign(y_i)}_{z_i} \\
        -y_i              & \WithProb~ \epsilon^{\sign(y_i)}_{z_i}.
    \end{cases}
\end{align*}
This reflects a setting where noise rates are independent of the $\bm{x}_i$ at fixed values $y_i$ and $z_i$ (e.g., a medical problem where $y_i$ is the presence of a disease, and the disease is diagnosed less reliably for females $z_i = 1$).

In this paper, we mainly focus on two specific fairness constraints: Equal Opportunity and Equal Odds~\cite{hardt2016equality}. Equal opportunity requires that each group achieves equal \emph{true positive rate} (TPR) or \emph{false positive rate} (FPR), while equal odds requires both equal TPR and equal FPR. We use the following shorthand to denote different measures of performance, including TPR and FPR, computed for each group using the true labels $y$ and the noisy labels $\noisy{y}$, where $y, \noisy{y}\in \Set{+1,-1}$: 
\begin{equation*}
\begin{split}
\TPR_z & := \Probability(f(X)=+1 \mid Y=+1, Z=z) \\
\FPR_z & := \Probability(f(X)=+1 \mid Y=-1, Z=z) \\
\widetilde{\TPR}_z & :=\Probability(f(X)=+1 \mid \noisy{Y}=+1, Z=z) \\ \widetilde{\FPR}_z & := \Probability(f(X)=+1 \mid \noisy{Y}=-1, Z=z)
\end{split}
\end{equation*}
$\widetilde{\TPR}_z$ and $\widetilde{\FPR}_z$ are taken with respect to the noisy labels.
\section{Enforcing Fairness Constraints on Noisy Labels Can be Harmful}\label{sec:Analysis}
Recent results have established that enforcing fairness constraints improves classifier accuracy when the labels suffer from label noise that is \emph{uniform} across different groups~\cite{Blum2019RecoveringFB}. However, as we shall see, adding fairness constraints can lead to harm when \emph{group-dependent} noise is present in the labels.

\subsection{Parity Constraints on Noisy Labels Harms Groups with Clean Labels}

The first message that we wish to deliver is that \emph{naively enforcing parity constraints on the noisy labels may harm the accuracy of the classifier for the groups that are not affected by label noise}. Without loss of generality, we present our results in settings where we wish to train a classifier with equal $\TPR$ across groups. Similar derivations hold for other related constraints (e.g., the ones as linear combinations of the entries in the confusion matrix), such as equal FPR, and equal balance error (BER) \cite{menon2018cost}. 

Consider a classification problem with two identical groups $z$ and $z'$ where samples from group $z$ have uncorrupted labels while samples from group $z'$ have noisy labels. On the clean data, the parity constraints naturally hold since the data for both groups is drawn from an identical distribution. We next show that the label noise presented in group $z'$ can harm the clean group $z$ when enforcing parity constraints. Formally:

\begin{theorem}\label{thm:NoiseHarmsAccuracy}
Consider a setting with two identical groups $(X,Y,Z=z)$ and $(X,Y,Z=z')$.
Group $z$ has clean labels, i.e., $\epsilon^+_z = \epsilon^-_{z} = 0$. Group
$z'$ suffers from symmetric noise $\epsilon^+_{z'} = \epsilon^-_{z'} = e > 0$. In this setting, a classifier trained subject to the equal TPR constraint ($\TPR_z = \widetilde{\TPR}_{z'}$) leads to an uninformative classifier that $\TPR_z = \FPR_z$. 
\end{theorem} 

We defer the proof to Section Ommited Proofs. Thus, even if group $z$ is represented with completely uncorrupted labels in the training data, the imposition of equal TPR in the presence of noise for $z'$ will diminish the classifier's predictive accuracy on members of group $z$.

\paragraph{Case study.}
    \begin{table}
        \centering
        \caption{Label noise harms accuracy: Adult dataset. High FPR implies weak discrimination power. We highlight any high harm the classifier suffers when enforcing equal TPR.}
        \label{tab:NoiseHarmsAccuracy}
        \begin{tabular}{l l c c c}
            \toprule
            Metrics & Groups & $f$ & & $f_{\textsf{fair}}$ \\
            \midrule
            \multirow{2}{*}{TPR} & \emph{female} & 97.12\% & $\Rightarrow$ &96.44\% \\ 
            & \emph{male} & 92.40\% &$\Rightarrow$ & 98.26\% \\\hline
            \multirow{2}{*}{FPR} & $female$ & 53.35\% &$\Rightarrow$ & \textcolor{ACMRed}{$78.11\%$} \\ 
            & \emph{male} & 46.81\% &$\Rightarrow$ & \textcolor{ACMRed}{$84.32\%$} \\\midrule
            \multirow{2}{*}{Accuracy} & $female$ & 91.62\% & $\Rightarrow$ & 88.32\% \\
            & \emph{male} & 80.39\% & $\Rightarrow$ & \textcolor{ACMRed}{$72.97\%$} \\
            \bottomrule
        \end{tabular}
    \end{table}
We empirically examine the above observation on the Adult dataset from UCI Machine Learning repository~\cite{Asuncion2007UCIML}. There are two sensitive groups, $Z = \{male, female\}$, in this data set. We inject symmetric noise $\epsilon^+ = \epsilon^- = 0.3$ into labels for members of the  $female$ group. Then, we train two classifiers: $f$, which is trained without any fairness constraints, and $f_{\textsf{fair}}$, which is trained with the imposition of equal TPR using the reduction method~\cite{agarwal2018reductions}. As is shown in Table~\ref{tab:NoiseHarmsAccuracy}, the empirical results mirror Theorem~\ref{thm:NoiseHarmsAccuracy}. When the difference between $f_{\textsf{fair}}$'s TPR for the two groups becomes small (less than 2\%), $f_{\textsf{fair}}$'s TPR and FPR become close together, and the accuracy decreases significantly. The above trends hold even when we try to equalize TPR and FPR together across groups. We notice that the two groups are not strictly identical in the Adult dataset, but our example implies that there exists dangerous cases where enforcing fairness constraints can harm classifier accuracy for the group with uncorrupted labels.

\subsection{Violation of Fairness under Perceived Fairness}
\label{Sec::TheoryHarmFromFairERM}

Our second message is that \textit{training fair classifiers using noisy labels may lead to a false impression of fairness}. This arises when the fairness constraints are satisfied over the noisy labels while being violated over the clean labels. Before proceeding, we require extending Proposition 16 of~\cite{menon2015learning} into the situation with group-dependent label noise. A similar result appears in~\cite{Scott2013ClassificationWA}.

\begin{lemma}\label{noisey:equalTPR}
For each group $z$ we have that
\begin{align}
\TPR_z = &(1-\epsilon^+_z) \cdot \widetilde{\TPR}_z+ \epsilon^+_z \cdot \widetilde{\FPR}_z \\ \FPR_z =  &\epsilon^-_z \cdot \widetilde{\TPR}_z+ (1-\epsilon^-_z) \cdot \widetilde{\FPR}_z 
\end{align}
\begin{proof}
Expanding $\Probability(f(X)=+1 \mid Y=+1, Z=z)$ using law of total probability we have
\begin{small}
\begin{align}
\TPR_z  &= \Probability(f(X)=+1\mid Y=+1, Z=z) \nonumber\\ 
    & = \Probability(f(X)=+1, \tilde{Y} = +1 \mid Y=+1, Z=z) \nonumber \\
    &\ + \Probability(f(X)=+1, \tilde{Y} = -1 \mid Y=+1, Z=z) \nonumber \\
    & = \Probability(\tilde{Y}=+1|Y=+1, Z=z) \cdot \Probability(f(X)=+1\mid \tilde{Y}=+1, Y=+1, Z=z) \nonumber \\
    &\ + \Probability(\tilde{Y}=-1|Y=+1, Z=z)\cdot \Probability(f(X)=+1\mid \tilde{Y}=-1, Y=+1, Z=z) \nonumber \\
    & = \Probability(\tilde{Y}=+1|Y=+1, Z=z) \cdot \Probability(f(X)=+1\mid \tilde{Y}=+1, Z=z) \nonumber \\
    &\ + \Probability(\tilde{Y}=-1|Y=+1, Z=z)\cdot \Probability(f(X)=+1\mid \tilde{Y}=-1, Z=z) \nonumber \\
    & = (1-\epsilon^+_z) \cdot \widetilde{\TPR}_z +\epsilon^+_z \cdot \widetilde{\FPR}_z
\end{align}
\end{small}
Note in the above we drop the dependence on $Y$ when conditioning on $\tilde{Y}$. This is because $f$ is trained purely on the noisy labels, and $\tilde{Y}$ encodes all the information $f$ has about $Y$.

A similar derivation holds for $\FPR_z$.
\end{proof}
\end{lemma}
We also note that, in the special case where all groups suffer from an identical rate of label corruption, the learner \emph{can} be oblivious to the specific error rates:

\begin{theorem}\label{thm:NoiseDoesNotHarmFairness}
Consider a classification problem with noisy labels where the noise rates are independent of group membership, so that $\epsilon^+_z=\epsilon^+_{z'}$ and $\epsilon^-_z=\epsilon^-_{z'}~ \forall z,z'\in Z$. Then it follows that $\TPR_z =\TPR_{z'}~ \forall z,z' \in Z$, if equal odds (equalizing both $\TPR$ and $\FPR$) on the noisy labels is imposed.
\end{theorem}

The proof follows by applying the assumption of equal error rates and equal odds on the noisy labels with Lemma~\ref{noisey:equalTPR}. However, things break down in the general case. If we impose equal odds across groups on a learner that is unaware of the labels' noisiness (i.e. whenever $\widetilde{\TPR}_z=\widetilde{\TPR}_{z'}$), then:

\begin{theorem}\label{thm:NoiseHarmsFairness}
Assume that a classifier is subject to equal odds in the presence of group-dependent label noise. Then for any two groups $z, z' \in Z$, we have
\begin{align*}
|\TPR_z-\TPR_{z'}| & = |\widetilde{\TPR}_z-\widetilde{\FPR}_z|\cdot |\epsilon^+_z - \epsilon^+_{z'}|, \\
|\FPR_z-\FPR_{z'}| & = |\widetilde{\TPR}_z-\widetilde{\FPR}_z|\cdot |\epsilon^-_z - \epsilon^-_{z'}|.
\end{align*}
Unless the classifier is random on the noisy training data, i.e., $\widetilde{\TPR}_z =\widetilde{\FPR}_{z}$, it is impossible to satisfy equal odds over the clean data whenever $\epsilon^+_z \ne \epsilon^+_{z'}$ and $\epsilon^-_z \neq \epsilon^-_{z'}$.
\begin{proof}
    Noticing that $\widetilde{\TPR}_z = \widetilde{\TPR}_{z'}$ and $\widetilde{\FPR}_z = \widetilde{\FPR}_{z'}$ (equalizing fairness metrics on the noisy data) and applying Lemma~\ref{noisey:equalTPR}, we obtain
    \begin{small}
    \begin{align*}
        |\TPR_z - \TPR_{z'}| & = |((1-\epsilon_z^+) \cdot \widetilde{\TPR}_z + \epsilon_z^+ \cdot \widetilde{\FPR}_z) \\
        & \quad -  ((1-\epsilon_{z'}^+) \cdot \widetilde{\TPR}_{z'} + \epsilon_{z'}^+ \cdot \widetilde{\FPR}_{z'})| \\
        & = |\epsilon_z^+ \cdot (\widetilde{\FPR}_z - \widetilde{\TPR}_z) - \epsilon_{z'}^+ \cdot (\widetilde{\FPR}_{z} - \widetilde{\TPR}_{z})| \\
        & = |(\epsilon_{z}^+ - \epsilon_{z'}^+) \cdot (\widetilde{\FPR}_{z} - \widetilde{\TPR}_{z})| \\
        & = |\widetilde{\TPR}_{z} - \widetilde{\FPR}_{z}| \cdot |\epsilon_{z}^+-\epsilon_{z'}^+|
    \end{align*}
    \end{small}
The argument for $\FPR$ is symmetrical:
    \begin{small}
    \begin{align*}
        |\FPR_z - \FPR_{z'}| & = |(\epsilon_z^- \cdot \widetilde{\TPR}_z + (1-\epsilon_z^-) \cdot \widetilde{\FPR}_z) \\
        &\quad - (\epsilon_{z'}^- \cdot \widetilde{\TPR}_{z'} + (1-\epsilon_{z'}^+) \cdot \widetilde{\FPR}_{z'})| \\
        & = |\epsilon_z^- \cdot (\widetilde{\TPR}_z - \widetilde{\FPR}_z) - \epsilon_{z'}^- \cdot (\widetilde{\TPR}_{z} - \widetilde{\FPR}_{z})| \\
        & = |(\epsilon_{z}^- - \epsilon_{z'}^-) \cdot (\widetilde{\TPR}_{z} - \widetilde{\FPR}_{z})| \\
        & = |\widetilde{\TPR}_{z} - \widetilde{\FPR}_{z}| \cdot |\epsilon_{z}^- -\epsilon_{z'}^-|
    \end{align*}
    \end{small}
    Therefore 
    \[
      |\TPR_z - \TPR_{z'}| > 0,    |\FPR_z - \FPR_{z'}|> 0,
    \]
    when $\widetilde{\TPR}_{z} \neq \widetilde{\FPR}_{z}$, $\epsilon_{z}^+ \neq \epsilon_{z'}^+, \epsilon_{z}^- \neq \epsilon_{z'}^-$.
\end{proof}
\end{theorem}
The proof follows by a direct application of Lemma~\ref{noisey:equalTPR}. Theorem~\ref{thm:NoiseHarmsFairness} implies that the true fairness violation is proportional to the difference in error rates across the different sub-groups. We offer two remarks. First, if the error rates are systematically biased towards a particular group, then a perceived fair classifier will lead to unequal odds. Second, the above bias will be reinforced when the trained model is more accurate on noisy data; a more accurate model will lead to a larger difference in $|\widetilde{\TPR}_z-\widetilde{\FPR}_z|$.

\section{Fair ERM with Noisy Labels}\label{sec:fairERM}
In this section, we describe two noise-tolerant and fair ERM solutions that address the combined challenges of heterogeneous and group-dependent label noise. Both the surrogate loss and group-weighted peer loss approaches for handling noisy labels rely on estimations of the label noise. Our procedure for estimating the noise parameters, detailed in Section~\ref{sec:estimation}, is an adaptation of~\cite{Northcutt2019ConfidentLE}.  Section~\ref{sec:estimation} also offers discussion of the impacts of noisy estimates.

\begin{table}[bb]
\caption{Surrogate constraints for surrogate loss.}
\label{table:surrogateconstraint}
\begin{tabular}{lc}
\toprule
Metric &  $\reallywidehat{F}_z(f)$\\ 
\midrule
$\TPR$ & $ (1-\epsilon^+_z) \cdot \reallywidehat{\TPR}_z+ \epsilon^+_z \cdot \reallywidehat{\FPR}_z $ \\ 
$\FPR$ & $\epsilon^-_z \cdot \reallywidehat{\TPR}_z+ (1-\epsilon^-_z) \cdot \reallywidehat{\FPR}_z$  \\
Equal Odds & both TPR and FPR  \\  
\bottomrule
\end{tabular}
\end{table}

\begin{table}[t]
\centering
\caption{Surrogate constraints for group weighted peer loss}
\label{tab:gp_constraint}
\begin{tabular}{lc}
\toprule 
Metric &  $\reallywidehat{F}_z(f)$\\ 
\midrule
$\TPR$ & $ \Probability(f(X)=+1|Z=z) + \frac{\Delta_z}{2} (\reallywidehat{\TPR}_z - \reallywidehat{\FPR}_z) $ \\ 
$\FPR$ & $\Probability(f(X)=+1|Z=z) - \frac{\Delta_z}{2} (\reallywidehat{\TPR}_z - \reallywidehat{\FPR}_z)$  \\
Equal Odds & both TPR and FPR  \\  
\bottomrule
\end{tabular}
\end{table}

\subsection{A Surrogate Loss Approach}\label{sec:SurrogateLoss}
As we shall see, training an unmodified loss function using the noisy labels $\noisy{y}_i$ corrupts the model in a manner that cannot be addressed via post-hoc correction. Thus, a natural resolution is to modify the loss function itself. This modified loss function is called a \textit{surrogate loss}.

\subsubsection*{Bias removal surrogate loss functions.} Bias removal via a surrogate loss is a popular approach to handling label noise~\cite{natarajan2013learning}. The original loss function $\ell(\cdot)$ is replaced with a surrogate loss function $\noisy{\ell}(\cdot)$ that ``corrects" for noise in the labels in expectation. Formally, the surrogate loss is chosen so that the cost of mis-classifying an element $\bm{x}_i$ with true label $y_i$ is equivalent to the expected loss value that arises from using noisy label $\noisy{y}_i$. Thus, we want to find a surrogate loss $\noisy{\ell}$ such that:
\begin{align}
\ell(f(\bm{x}),y)=\Expectation_{\noisy{Y}}[\noisy{\ell}(f(\bm{x}),\noisy{Y})\mid Y=y] \label{eq:SurrogateLossCondition}
\end{align}
for all $\bm{x}$ and $y$. When the noise depends on the label value, the function given by

\begin{small}
\begin{align}
\noisy{\ell}(f(\bm{x}_i),\noisy{y}_i=+1) &:= 
    \frac
    {(1-\epsilon^-_{z_i})\ell(f(\bm{x}_i),+1)-\epsilon^+_{z_i}\ell(f(\bm{x}_i),-1)}
    {1-\epsilon^+_{z_i}-\epsilon^-_{z_i}}, \label{surrogate:pos} \\
\noisy{\ell}(f(\bm{x}_i),\noisy{y}_i=-1) &:=
    \frac
    {(1-\epsilon^+_{z_i}) \ell(f(\bm{x}_i),-1)-\epsilon^-_{z_i} \ell(f(\bm{x}_i),+1)}
    {1-\epsilon^+_{z_i}-\epsilon^+_{z_i}}. \label{surrogate:neg}
\end{align}
\end{small}
satisfies the above property, as shown by Lemma 1 in~\cite{natarajan2013learning}. A classifier $f$ minimizing the surrogate loss on noisy data $\tilde{\ell}(X, \tilde{Y})$ will minimize the loss on clean data $\ell(X, Y)$ in expectation. This property allows us to perform model selection on a noisy validation set, and one could choose the model that performs better on the validation set to deploy.

\subsubsection*{Surrogate fairness constraints.}
We will also need to modify the fairness constraints to account for the effects of noise. Our method of doing so is inspired by the surrogate loss that we need to work with an unbiased estimate of the fairness constraints. For the case of binary classification, we can express the surrogate measures of group-based fairness constraints using Lemma~\ref{noisey:equalTPR}.

We use Equation~(\ref{surrogate:pos}) and Equation~(\ref{surrogate:neg}) to define our surrogate loss functions $\noisy{\ell}_z(f(\bm{x}_i),\noisy{y}_i=+1)$, and $\noisy{\ell}_z(f(\bm{x}_i),\noisy{y}_i=-1)$. Furthermore, define the empirical TPR and FPR over the noisy labels as follows:
\begin{align}
    \reallywidehat{\TPR}_z(f) & = 
        \frac
            {\#(f(\bm{x}_i)=+1, \noisy{y}_i=+1, z_i=z)}
            {\#(\noisy{y}_i=+1, z_i=z)} \\  
        \reallywidehat{\FPR}_z(f) &  =
        \frac
            {\#(f(\bm{x}_i)=+1, \noisy{y}_i=-1, z_i = z)}
            {\# (\noisy{y}_i=-1, z_i=z)}
\end{align}
We then define our surrogate fairness measures $\reallywidehat{F}_z(f)$ using only noisy data, as detailed in Table \ref{table:surrogateconstraint}. Our noise-resistant fair ERM states as follows:
\begin{align}\label{eqn:noisyERM}
\min_{f\in\Hypotheses} &~ \sum_{i=1}^n \noisy{\ell}(f(\bm{x}_i),\noisy{y}_i)~~~~\notag \\ 
\SuchThat &~~~~ |\reallywidehat{F}_z(f) - \reallywidehat{F}_{z'}(f)| \le \delta, ~\forall z,z'.  
\end{align}

\subsection{Group Weighted Peer Loss Approach}\label{sec:PeerLos}

The recently developed \emph{peer loss} function partially circumvents the issue of noise rate estimation~\cite{Liu_2019}. The peer loss requires less prior knowledge of the noise rates for each class. It is defined as:
\begin{equation}\label{eq:peer_loss}
    \ell_{peer}(f(\bm{x}_i),\tilde{y}_i) := \ell\left(f(\bm{x}_i),\tilde{y}_i\right) - \alpha \cdot 
    \ell\left(f(\bm{x}_{i_1}),\tilde{y}_{i_2}\right),~
\end{equation}
where \[
\alpha = 1 - (1 - \epsilon^- - \epsilon^+) \cdot \frac{\Probability(Y=+1)-\Probability(Y=-1)}{\Probability(\tilde{Y}=+1) - \Probability(\tilde{Y}=-1)}
\] 
is a parameter to balance the instances for each label, and where $i_1$ and $i_2$ are uniformly and randomly selected samples from $I_z / \Set{i}$ (i.e., ``peer" samples which inspired the name peer loss as noted in \cite{Liu_2019}). Although the noise parameters explicitly appear in the definition of $\alpha$, only the knowledge of $\Delta := 1 - \epsilon^- - \epsilon^+$ is needed. In practice, we could tune $\alpha$ as a hyper-parameter during training. This loss function has the following important property, proven in Lemma 3 of~\cite{Liu_2019}:
\begin{equation}\label{lemma:pl}
    \Expectation_{\tilde{\mathcal{D}}_z}[\ell_{peer}(f(X), \tilde{Y})] = \Delta_z \cdot \Expectation_{\mathcal{D}_z}[\ell_{peer}(f(X), Y)],
\end{equation}
where $\tilde{\mathcal{D}}_z$ denotes the noisy distribution for group $z$ and $\Delta_z=1-\epsilon^-_z-\epsilon^+_z$. Adapting the peer loss function to labels with group dependent noise requires accounting for the differing values of $\Delta_z$. We do so by re-weighting Equation~(\ref{eq:peer_loss}) to obtain our \textit{group-weighted peer loss} $\ell_{gp}$:
\begin{equation}\label{eq:group-weighted-peer-loss}
    \ell_{gp}(f(\bm{x}_{i}),\tilde{y}_i) := \frac{1}{\Delta_{z_i}}\left(\ell(f(\bm{x}_i),\tilde{y}_i) - \alpha\cdot \ell\left(f(\bm{x}_{i_1}),\tilde{y}_{i_2}\right)\right).
\end{equation}
 When class is balanced for every group $z$, i.e., $\Probability_{Z=z}(Y=+1)=\Probability_{Z=z}(Y=-1)=\frac{1}{2}$, the parameter $\alpha$ is exactly $1$. In this case, the expected group-weighted peer loss on the noisy distribution $\noisy{\mathcal{D}}$ is the same as the expected uncorrected loss $\ell$ on the true distribution $\mathcal{D}$. More precisely:

\begin{theorem}\label{thm:group_weighted_peer_loss}
For all group dependent noise rates $\epsilon_z^-$ and $\epsilon_z^+$ satisfying $\epsilon_z^- + \epsilon_z^+ < 1$, taking $\ell(\cdot)$ as the 0-1 loss $ \Indicator(\cdot)$ and when $\Probability_{Z=z}(Y=+1)=\Probability_{Z=z}(Y=-1)=\frac{1}{2}$,
\begin{align}
    \Expectation_{\tilde{\mathcal{D}}}[\ell_{gp}(f(X), \tilde{Y})] = \Expectation_{\mathcal{D}}[\ell (f(X), Y] - \frac{1}{2}.
\end{align}
\begin{proof}
Observe that
    \[
        \ell_{gp}(f(\bm{x}_i), \tilde{y}) = \frac{1}{\Delta_{z_i}} \ell_{peer}(f(\bm{x}_i), \tilde{y})
    \]
Taking expectations over noisy data, we have
\begin{small}
    \begin{align*}
        &\Expectation_{\tilde{\mathcal{D}}}[\ell_{gp}(f(X), \tilde{Y})] \\
        & = \frac{1}{|I|} \cdot \sum_{z \in Z} |I_z| \cdot \Expectation_{\tilde{\mathcal{D}}_z}[\ell_{gp}(f(X_z), \tilde{Y}_z)] \\
        & = \frac{1}{|I|} \cdot \sum_{z \in Z} \frac{|I_z|}{\Delta_z} \cdot \Expectation_{\tilde{\mathcal{D}}_z}[\ell_{peer}(f(X_z), \tilde{Y}_z)] \\
        & = \frac{1}{|I|} \cdot \sum_{z \in Z} \frac{|I_z|}{\Delta_z} \cdot \Delta_z \Expectation_{\mathcal{D}_z}[\ell_{peer}(f(X_z), Y_z)] \tag*{(by Equation \ref{lemma:pl})}\\
        & = \Expectation_{\mathcal{D}}[\ell_{peer}(f(X), Y)] \numberthis \label{eq:expectation_gw_peer}
    \end{align*}
\end{small}
Notice that $\alpha=1$ when $\Probability(Y=+1) = \Probability(Y=-1) = \frac{1}{2}$, the definition of peer loss function gives
\begin{small}
\begin{align}\label{eq:expectation_pl}
    \Expectation_{X, Y}[\ell_{peer}(f(X), Y)] & = \Expectation_{X, Y}[\ell(f(X), Y)] - \Expectation_{X}\Expectation_{Y}[\ell(f(X), Y)]
\end{align}
\end{small}
Using the assumption that $\Probability(Y=+1) = \Probability(Y=-1) = \frac{1}{2}$ and the fact that $\ell$ is 0-1 loss function,
\begin{small}
\begin{align}
    \Expectation_{X}\Expectation_{Y}[\ell(f(X), Y)] & = \Probability(Y = +1) \cdot \Expectation_{X}[\ell(f(X), +1)] + \nonumber \\ & \quad \Probability(Y = -1) \cdot \Expectation_{X}[\ell(f(X), -1)] \nonumber \\ 
    & = \frac{1}{2}\cdot  \ell(f(X), +1) + \frac{1}{2}\cdot  \ell(f(X), +1) \nonumber  \\
    & = \frac{1}{2} \cdot \Indicator(f(X) \neq +1) + \frac{1}{2} \cdot \Indicator(f(X) \neq -1) \nonumber \\
    & = \frac{1}{2} \Probability(f(X) = -1) + \frac{1}{2}\cdot  \Probability(f(X) = +1) \nonumber \\
    & = \frac{1}{2} \label{eq:pl01loss}
\end{align}
\end{small}
Combining \cref{eq:expectation_gw_peer}, \cref{eq:expectation_pl} and \cref{eq:pl01loss}, we complete the proof 
\begin{small}
\begin{align*}
        \Expectation_{\tilde{\mathcal{D}}}[\ell_{gp}(f(X), \tilde{Y})] = \Expectation_{\mathcal{D}}[\ell (f(X), Y] - \frac{1}{2}
\end{align*}
\end{small}
\end{proof}
\end{theorem}

\subsubsection*{Peer-based surrogate fairness constraints.}
We acquire the following result in order to create group-aware surrogate constraints:
\begin{lemma}\label{lem:PeerLossTPRIdentities}
True $\TPR$ and $\FPR$ relate to $\widetilde{\TPR}_z, \widetilde{\FPR}_z$ defined on the noisy labels as follows:
\begin{small}
\begin{align}
    \TPR_z & = \Probability(f(X)=+1|Z=z) + \Delta_z \cdot (\widetilde{\TPR}_z - \widetilde{\FPR}_z) \cdot \Probability(Y=-1|Z=z) \\
    \FPR_z & = \Probability(f(X)=+1|Z=z) - \Delta_z \cdot  (\widetilde{\TPR}_z - \widetilde{\FPR}_z) \cdot \Probability(Y=+1|Z=z)
\end{align}
\end{small}
\begin{proof}
Following Lemma~\ref{noisey:equalTPR} we have,
\begin{small}
\begin{align*}
    \TPR_z - \FPR_z = (1 - \epsilon_z^+ - \epsilon_z^-) (\widetilde{\TPR} - \widetilde{\FPR}) = \Delta_z\cdot (\widetilde{\TPR} - \widetilde{\FPR})
\end{align*}
\end{small}
Notice that
\begin{small}
\begin{align*}
    \Probability(f(X)=+1 \mid Z=z) & = \Probability(Y=+1 \mid Z=z) \cdot \Probability(f(X)=+1 \mid Y = +1, Z=z) \\ & + \Probability(Y=-1|Z=z)\cdot \Probability(f(X) = +1 \mid Y=-1, Z=z)\\
    & =\Probability(Y=+1|Z=z) \cdot \TPR_z + \Probability(Y=-1|Z=z)\cdot \FPR_z \\
\end{align*}
\end{small}
Solving the two equations above we complete the proof.
\end{proof}
\end{lemma}
Lemma~\ref{lem:PeerLossTPRIdentities} allows us to derive the appropriate surrogate fairness constraints for the peer loss, displayed in Table~\ref{tab:gp_constraint}. Note that we have assumed that the datasest is balanced for each group; i.e., $\forall z \in Z \quad \Probability(Y=+1|Z=z) = \frac{1}{2}$. If the data is imbalanced, we will require knowing the marginal prior $\Probability(Y=+1|Z=z)$. We note that it is straightforward to get the estimated marginal priors as given by Equation (\ref{eq:marginal_priors}) in Section~\ref{sec:estimation}.

We merely require knowledge of $\Delta_z$ for each $z$ in order to define $\ell_{gp}$ and $\reallywidehat{F}_z(f)$. This is a weaker requirement compared to knowing the error rates (which will carry estimation of two parameters for each group). We indeed see our group peer loss approach performs more stably as compared to the surrogate loss approach introduced in last subsection when using noisy estimates of the noise rates. With group-weighted peer loss function and surrogate fairness constraints, we are able to perform a fair ERM as detailed in Equaltion (\ref{eqn:noisyERM}) by replacing $\tilde{\ell}$ with $\ell_{gp}$ and the corresponding $\reallywidehat{F}_z(f)$ term.

\subsection{Error Rates Estimation and its Impact}\label{sec:estimation}
We employ ``confident learning'' to perform noise rate estimation in our experiments~\cite{Northcutt2019ConfidentLE}. The first step is to pre-train a classifier $f_{pre}$ over the noisy labels directly and learn a noisy predicted probability $$
\hat{p}(y;\bm{x},z)=\Probability(f_{pre}(\bm{x})=y|Z=z).$$
Then, for each pair of classes $k,l \in \{+1, -1\}$, we define the subset of samples:
\begin{align*}
&\widehat{X}_{\hat{y}=k,z} := \{ \bm{x}_i | \tilde{y}_i = k, i \in I_z \}, \\ 
& \widehat{X}_{\hat{y}=k, y=l, z} := \{ \bm{x}_i | \tilde{y}_i = k, \hat{p}(y=l;\bm{x}_i, z) \geq t_{l,z}, i \in I_z\},
\end{align*}
where $$t_{l, z} = \frac{1}{|\widehat{X}_{\hat{y}=l,z}|} \sum_{\bm{x} \in \widehat{X}_{\hat{y}=l,z}} \hat{p}(\hat{y}=l;\bm{x}, z)$$ is the \emph{expected self-confidence probability} for class $l$ and group $z$. Using the above quantities, we estimate the group-aware joint probability $\widehat{Q}_{\tilde{y}=k, y=l, z} = \Probability(\widetilde{Y}=k, Y=l, Z=z)$ over the noisy labels  $\tilde{y}$ and clean labels $y$ with:
\begin{small}
\begin{equation}
    \widehat{Q}_{\tilde{y}=k, y=l, z} = \frac{
        \frac{|\widehat{X}_{\tilde{y}=k, y=l, z}|}{\sum_l |\widehat{X}_{\tilde{y}=k, y=l, z}|} \cdot |\widehat{X}_{\tilde{y} = k, z}|
    }{
        \sum_{k, l} \left(\frac{|\widehat{X}_{\tilde{y}=k, y=l, z}|}{\sum_l |\widehat{X}_{\tilde{y}=k, y=l, z}|} \cdot |\widehat{X}_{\tilde{y} = k, z}|\right)
    }
\end{equation}
\end{small}
We use the marginals of estimated joint to compute the noise parameter estimates for each group $z$:
\begin{small}
\begin{equation}\label{eq:estimate_epsilon}
    \begin{split}
        \hat{\epsilon}_z^+ & = \frac{\widehat{Q}_{\tilde{y}=-1,y=+1, z}}{\widehat{Q}_{\tilde{y}=-1,y=+1,z} + \widehat{Q}_{\tilde{y}=+1,y=+1,z}},\\
        \hat{\epsilon}_z^- & = \frac{\widehat{Q}_{\tilde{y}=+1,y=-1,z}}{\widehat{Q}_{\tilde{y}=+1,y=-1,z} + \widehat{Q}_{\tilde{y}=-1,y=-1,z}} \\
    \end{split}
\end{equation}
\end{small}
To estimate $\Delta_z$, we simply substitute $\hat{\epsilon}^-_z$ and $\hat{\epsilon}^+_z$ for $\epsilon^-_z$ and $\epsilon^+_z$ in the equation for $\Delta_z$. As a byproduct, we could estimate the marginal priors $\Probability(Y=+1|Z=z)$ by
\begin{small}
\begin{equation}\label{eq:marginal_priors}
\frac{\widehat{Q}_{\tilde{y}=+1,y=+1, z} + \widehat{Q}_{\tilde{y}=-1,y=+1, z}}{ \widehat{Q}_{\tilde{y}=+1,y=+1, z} + \widehat{Q}_{\tilde{y}=-1,y=+1, z} + \widehat{Q}_{\tilde{y}=-1,y=+1, z} + \widehat{Q}_{\tilde{y}=-1,y=-1, z}}
\end{equation}
\end{small}
\subsubsection*{Effects of noisy estimates.} It is important to quantify the impact of the noise rate estimation error on the accuracy and fairness of the resulting classifier. We first note that, for any $\eta,\tau>0$, the law of large numbers implies that taking sufficiently many samples from $\mathcal{D}$ will ensure that the following holds for all $z$ with probability at least $1-\eta$:
\begin{equation}
\begin{split}
    \max
    \Big\{& 
        \left|\noisyeps^+_z - \epsilon^+_z\right|,
        \textstyle{\left|\frac{\noisyeps^+_z}{1-\noisyeps^+_z-\noisyeps^-_z}- \frac{\epsilon^+_z}{1-\epsilon^+_z-\epsilon^-_z}\right|}, \\
              & \left|\noisyeps^-_z-\epsilon^-_z\right|,
        \textstyle{\left|\frac{1-\noisyeps^-_z}{1-\noisyeps^+_z-\noisyeps^-_z}- \frac{1-\epsilon^-_z}{1-\epsilon^+_z-\epsilon^-_z}
        \right|} 
    \Big\}
  \le \tau.
  \end{split}\label{eq:NoiseBounds}
\end{equation}

Denote by $\hat{\ell}(\cdot)$ the surrogate loss function defined using the estimated noises $\Set{\noisyeps^+_z, \noisyeps^-_z}$, and let
\begin{small}\begin{align*}
    \hat{f}^* = \argmin_{f \in\Hypotheses} \sum_{i=1}^N \hat{\ell}(f(\bm{x}_i),\tilde{y}_i), \qquad
    \tilde{f}^* = \argmin_{f \in \Hypotheses} \sum_{i=1}^N \tilde{\ell}(f(\bm{x}_i),\tilde{y}_i)
\end{align*}\end{small}
We have the following result and defer the proof to Section~\ref{sec:proofs}:
\begin{theorem}\label{thm:effect:noise:e}
For every $\eta,\tau>0$ there exists $N$ such that
\begin{small}\begin{equation}
\frac{1}{N} \cdot \sum_{i=1}^N \noisy{\ell}(\hat{f}^*(\bm{x}_i),\noisy{y}_i)- \frac{1}{N} \cdot  \sum_{i=1}^N \noisy{\ell}(\tilde{f}^*(\bm{x}_i),\noisy{y}_i) \leq 4\tau \cdot \bar{\ell}
\end{equation}\end{small}
with probability at least $1-\eta$, where $\bar{\ell} := \max \ell$.
\end{theorem}
Because the fairness constraints $\reallywidehat{F}_z(f)$ are linear in $\epsilon^+_z,\epsilon^-_z$s, the additional fairness violations incurred due to the noisy estimates of the error rates will also be linear in $\tau$ too. Similar observations hold when using the estimated $\tilde{\Delta}_z$ in the peer loss.
\section{Experiments}
\label{Sec::Experiments}

\begin{table*}[htb]
    \centering
    \caption{Dataset statistic and parameters.}
    \label{tab:dataset_statistic}
    \resizebox{0.7\linewidth}{!}{\begin{tabular}{c l c c l c c}
    \toprule
        \multirow{2}{*}{Dataset} & \multirow{2}{*}{Source} & \multirow{2}{*}{Number of data examples $n$} & \multirow{2}{*}{Fairness Tolerance $\delta$} & \multirow{2}{*}{Sensitive Groups} & \multicolumn{2}{c}{Noise Rates}  \\
        & & & & & $\epsilon^-$ & $\epsilon^+$ \\
    \midrule
        \multirow{2}{*}{\texttt{adult}} & \multirow{2}{*}{UCI~\cite{Asuncion2007UCIML}} &\multirow{2}{*}{32561} & \multirow{2}{*}{$2\%$} & \emph{female} & 0.45 & 0.15 \\
        & & & & \emph{male} & 0.35 & 0.55 \\
    \midrule
        \multirow{2}{*}{\texttt{arrest}} & \multirow{2}{*}{COMPAS~\cite{angwin2016machinebias}} &\multirow{2}{*}{6644} & \multirow{2}{*}{$5\%$} & \emph{white} & 0.40 & 0.30 \\
        & & & & \emph{black} & 0.15 & 0.25 \\
    \midrule
        \multirow{4}{*}{\texttt{arrest}} & \multirow{4}{*}{COMPAS~\cite{angwin2016machinebias}} &\multirow{4}{*}{6644} & \multirow{4}{*}{$5\%$} & \emph{white} male & 0.45 & 0.10 \\
        & & & & \emph{black male} & 0.10 & 0.35 \\
        & & & & \emph{white female} & 0.35 & 0.45 \\
        & & & & \emph{black female} & 0.55 & 0.25 \\
    \midrule
        \multirow{4}{*}{\texttt{violent}} & \multirow{4}{*}{COMPAS~\cite{angwin2016machinebias}} &\multirow{4}{*}{5278} & \multirow{4}{*}{$5\%$} & \emph{white male} & 0.45 & 0.10 \\
        & & & & \emph{black male} & 0.10 & 0.35 \\
        & & & & \emph{white female} & 0.35 & 0.45 \\
        & & & & \emph{black female} & 0.55 & 0.25 \\
    \midrule
        \multirow{2}{*}{\texttt{German}} & \multirow{2}{*}{UCI~\cite{Asuncion2007UCIML}} &\multirow{2}{*}{1000} & \multirow{2}{*}{$2\%$} & \emph{female} & 0.45 & 0.15 \\
        & & & & \emph{male} & 0.35 & 0.55 \\   
    \midrule
        \multirow{2}{*}{\texttt{law}} & \multirow{2}{*}{LSAC~\cite{Wightman1998LSACNL}} &\multirow{2}{*}{18692} & \multirow{2}{*}{$2\%$} & \emph{white} & 0.45 & 0.15 \\
        & & & & \emph{black} & 0.35 & 0.55 \\ 
    \bottomrule
    \end{tabular}}
\end{table*}

Due to the difficulty of acquiring real world datasets with known label corruption characteristics, we artificially synthesize the datasets with a noise generation step. These controlled experiments help us understand the robustness of our approaches under different noise scenarios.

\subsection{Experimental Setup}
\paragraph{Dataset}
We evaluate our methods as well as other baseline methods on five datasets: 
\squishlist
\item \texttt{Adult}, the Adult dataset from the UCI ML Repository with males and females as the protected groups~\cite{Asuncion2007UCIML}.
\item \texttt{Arrest} and \texttt{Violent}, the COMPAS recidivism dataset for arrest and violent crime statistics, with race (restricted to white and black) and gender as the sensitive attributes~\cite{angwin2016machinebias}.
\item \texttt{German}, the German credit dataset from UCI ML Repository with gender as the sensitive attribute~\cite{Asuncion2007UCIML}.
\item \texttt{Law}, a subset of the original data set from LSAC with race (restricted to black and white) as the sensitive attribute~\cite{Wightman1998LSACNL}.
\squishend

Table~\ref{tab:dataset_statistic} describes the dataset statistics and parameters used in the experiments. We chose to apply a diverse set of noise parameters to the different subgroups. The fairness tolerance $\delta$ and noise parameters $\epsilon$ for \texttt{Adult}, \texttt{German} and \texttt{Law} data sets are identical, but they are different from \texttt{Arrest} and \texttt{Violent} data sets because \texttt{Arrest} and \texttt{Violent} data sets contain more protected groups. 
We make this choice mainly for the baseline models to obtain meaningful results to compare with.

\paragraph{Noise generation}
We randomly split the clean dataset $\mathcal{D} = \{(x_i, y_i, z_i)\}_{i=1}^n$ into a training set and a test set in a ratio of $80$ to $20$. We add asymmetric label noises to the training dataset, and leave the test data untouched for verification purposes. For each sensitive group $z\in Z$, we randomly flip the clean label $y$ with probability $\epsilon_z^-$ if its value is $-1$, and we flip the clean label with probability $\epsilon_z^+$ if it's $+1$. After injecting this noise, we use the same training set and test set to benchmark all the methods.

\paragraph{Methods.}
For all of the methods above, we use logistic regression to perform classification and leverage the reduction approach as proposed in~\cite{agarwal2018reductions} for solving our constrained optimization problem. We evaluate the performance of several methods: 
\squishlist
\item \texttt{Clean}, in which the classifier is trained on the clean data subject to the equal odds constraint 
\item \texttt{Corrupt}, which directly trains the classifier on the corrupted data subject to the equal odds fairness constraint 
\item \texttt{Surrogate Loss}, which uses the surrogate loss approach described in Section~\ref{sec:SurrogateLoss} 
\item \texttt{Group Peer Loss}, which uses the group weighted peer loss approach described in Section~\ref{sec:PeerLos} to train a fair classifier on the corrupted training set.
\squishend
The \texttt{Corrupt} baseline gives us a sense about the harm caused by the unawareness of the labels' noise, and the \texttt{clean} baseline shows the biases contained in the datasets.

 We set the same maximum fairness violation $\delta$ for all the methods on the same dataset during training. As there are more sensitive groups on \texttt{arrest} and \texttt{violent} datasets, we set $\delta = 5\%$ on these datasets and $\delta = 2\%$ on the other datasets. We report metrics for each of the above methods averaged over five runs.

\paragraph{Computing Infrastructure}
We conducted all the experiments on a 3 GHz 6-Core Intel Core i5 CPU. The running time for \texttt{Surrogate Loss} is about 10 minutes, while the running time for \texttt{Group Peer Loss} could be over 30 minutes.

\paragraph{Tuning $\alpha$ in Peer Loss}
The performance of our group weighted peer loss is highly influenced by the hyperparameter $\alpha$. Recall that
\[
    \Expectation_{\tilde{\mathcal{D}}_z}[\ell_{gp}(f(X), \tilde{Y})] = \Expectation_{\mathcal{D}_z}[\ell_{gp}(f(X), Y)]
\]
We split $10\%$ of data examples in the train set for validation and found the optimal $\alpha$ using grid search. The range of $\alpha$ we searched varied between $0.0$ to $2.0$. We observed that both the accuracy and fairness violation on the validation set exhibit the same trends on the test set. In practice, the group weighted peer loss with $\alpha = 0.3$ achieves the best performance on the \texttt{Adult} dataset.

\definecolor{best}{HTML}{BAFFCD}
\definecolor{issue}{HTML}{FFC8BA}
\definecolor{bad}{HTML}{FFC8BA}

\newcommand{\good}[1]{\cellcolor{best}#1} 
\newcommand{\bad}[1]{\cellcolor{issue}#1} 
\newcommand{\violation}[1]{\cellcolor{bad}#1} 

\begin{table*}[!tb]
\normalsize 
\caption{Overview of group-based performance metrics for all methods on 5 data sets. We highlight the best values achieved for fairness violation and accuracy in green and the worst in red. $m$ is the number of sensitive groups, $\bar{\epsilon}$ is the average of error rates over all the groups and all label classes $\epsilon^+_z,\epsilon^-_z$s. \textit{true} indicates training with true noise parameters and \textit{estimated} indicates training with estimated noise parameters. The values after $\pm$ are the standard deviation.}
\newcommand{\titlecell}[2]{\setlength{\tabcolsep}{0pt}{\small{\textsc{\begin{tabular}{#1}#2\end{tabular}}}}}
\newcommand{\bfcell}[2]{\setlength{\tabcolsep}{0pt}\textbf{\begin{tabular}{#1}#2\end{tabular}}}
\newcommand{\metrics}[0]{{\cell{l}{\textit{violation} \\ \textit{accuracy}}}}
\centering\resizebox{1\linewidth}{!}{\begin{tabular}{clccccccc}
\toprule
\normalsize 
&
& 
&
&
& \multicolumn{2}{c}{\textsc{Surrogate Loss}}
& \multicolumn{2}{c}{\textsc{Group Peer Loss}} 
\\
\cmidrule(lr){6-7} 
\cmidrule(lr){8-9}

\bfcell{c}{Dataset}
& \bfcell{l}{Metrics}
 & \bfcell{c}{Avg. $\bar{\epsilon}$}
 & \titlecell{c}{Clean}
 & \titlecell{c}{Corrupt}
 & \titlecell{c}{\textit{true}}
 & \titlecell{c}{\textit{estimated}}
 & \titlecell{c}{\textit{true}}
 & \titlecell{c}{\textit{estimated}}
 \\ 
  \toprule
\cell{c}{\textds{Adult}\\ $m=2$}
& \metrics{}
 & \cell{c}{0.38}
 & \cell{c}{0.47\% \\ $83.76$\%}
 & \cell{c}{\bad{$8.36 \pm 1.36$\%} \\ $76.08\pm2.49$\%}
 & \cell{c}{$1.46\pm0.50$\% \\ \good{$81.16\pm3.41$\%}}
 & \cell{c}{$1.39 \pm 0.80$\% \\ $75.99\pm7.45$\%}
 & \cell{c}{\good{$1.18\pm0.63$\%} \\ $77.00\pm2.52$\%}
 & \cell{c}{$1.69\pm0.86$\% \\ \bad{$75.13\pm5.15$\%}}
 \\ 
   
\midrule

\cell{c}{\textds{Arrest}\\$m=2$\\}
 & \metrics{}
 & \cell{c}{0.28}
 & \cell{c}{$2.27$\% \\ 65.16\%}
 & \cell{c}{\bad{$2.98\pm0.74$\%} \\ \bad{$60.72\pm0.66$\%}}
 & \cell{c}{$0.54\pm0.27$\% \\ $61.7\pm3.23$\%}
 & \cell{c}{\good{$0.36\pm 0.24$\%} \\ $62.3\pm5.30$\%}
 & \cell{c}{$1.78\pm0.89$\% \\ $63.81\pm3.35$\%}
 & \cell{c}{$1.05\pm0.55$\% \\ \good{$65.31\pm3.41$\%}}
 \\ 

 \\
\cell{c}{\textds{Arrest}\\$m=4$\\}
 & \metrics{}
 & \cell{c}{0.34}
 & \cell{c}{5.89\% \\ 66.0\%}
 & \cell{c}{\bad{$12.93\pm0.95$\%} \\ \bad{$53.7\pm1.82$\%}}
 & \cell{c}{\good{$0.88\pm0.27$\%} \\ \good{$65.7\pm2.92$\%}}
 & \cell{c}{$2.48\pm1.42$\% \\ $58.8\pm4.96$\%}
 & \cell{c}{$1.36\pm0.69$\% \\ $60.27\pm2.90$\%}
 & \cell{c}{$1.40\pm0.36$\% \\ $57.56\pm2.96$\%}
 \\
 
\midrule

\cell{c}{\textds{Violent}\\$m=4$}
 & \metrics{}
 & \cell{c}{0.34}
 & \cell{c}{0.37\% \\ 60.18\%}
 & \cell{c}{$7.16\pm0.80$\% \\ \bad{$52.2\pm0.23$\%}}
 & \cell{c}{$4.81\pm0.70$\% \\ $53.14\pm4.91$\%}
 & \cell{c}{\bad{$7.76\pm1.02$\%} \\ $55.4\pm0.71$\%}
 & \cell{c}{$2.06\pm0.81$\% \\ \good{$55.64\pm4.88$\%}}
 & \cell{c}{\good{$0.68\pm0.28$\%} \\ $52.7\pm0.57$\%}
 \\ 
   
\midrule

\cell{c}{\textds{German}\\$m=2$}
 & \metrics{}
 & \cell{c}{0.38}
 & \cell{c}{0.68\% \\ 74.5\%}
 & \cell{c}{$2.68\pm0.32$\% \\ $70.5\pm0.00$\%}
 & \cell{c}{\bad{$11.79\pm3.87$\%} \\ \bad{$68.5\pm4.27$\%}}
 & \cell{c}{$11.08\pm2.16$\% \\ \good{$71.5\pm2.53$\%}}
 & \cell{c}{\good{$0.00\pm0.00$\%} \\ $70.0\pm0.71$\%}
 & \cell{c}{$1.64\pm0.32$\% \\ $70.5\pm2.53$\%}
 \\ 
   
\midrule 

\cell{c}{\textds{Law}\\$m=2$}
& \metrics{}
 & \cell{c}{0.38}
 & \cell{c}{0.6\% \\ 90.67\%}
 & \cell{c}{\bad{$2.74\pm0.12$\%} \\ $90.16\pm0.79$\%}
 & \cell{c}{$0.36\pm0.08$\% \\ $90.26\pm0.48$\%}
 & \cell{c}{$1.98\pm1.16$\% \\ \bad{$89.92\pm2.86$}\%}
 & \cell{c}{\good{$0.03\pm0.02$}\% \\ \good{$90.32\pm0.10$\%}}
 & \cell{c}{$0.57\pm0.12$\% \\ $90.29\pm0.20$\%}
 \\ 
     
\bottomrule
\end{tabular}
}
\label{table:results}
\end{table*}

\subsection{Results}
\label{Sec::ExperimentalResults}
We present an overview of the performance for each method on the test set in Table~\ref{table:results}. We compare the two fair ERM approaches using both the true and estimated noise rates. The metrics we report include \textit{violation}, the maximum difference in TPR and FPR between groups $z, z' \in Z$, and \textit{accuracy}, the accuracy achieved on test set. 

We make the following observations about our results. First, both of the two fair ERM approaches in Section~\ref{sec:fairERM} produce classifiers that are more effective at mitigating unfairness than a classifier that is naively trained on the corrupted data.

In particular, the group weighted peer loss approach achieves almost $0\%$ violation on the \texttt{German} and \texttt{law} data sets, when given the true noise parameters. The only noticeable worse case arises when applying the surrogate loss approach to the \texttt{German} dataset. This may be due to the high variance of the \texttt{German} dataset, which has fewer than 1000 samples.

Second, as expected, models trained using our proposed fair ERM methods do not achieve the same level of accuracy as a model that is fit using clean labels. However, our models are typically more accurate than the model fit directly to the corrupted data. For example, on the \texttt{arrest} data set with four protected groups, the surrogate loss approach achieves a similar accuracy to the classifier trained on clean data while incurring an even smaller fairness violation. Third, Our methods perform similarly well when trained using both the true and with the estimated noise parameters, indicating that the noise estimation procedures are effective. On \texttt{arrest} and \texttt{violent} datasets, our methods with estimated noise parameters even perform better than those with true parameters. This is probably due to the biases and noise in these datasets. Finally, our fair ERM frameworks adapt well to multiple sensitive groups, as demonstrated by the good performance on the \texttt{Arrest} and \texttt{Violent} data sets.

\begin{table}[t]
    \centering
        \caption{We show how different levels of symmetric noise $\epsilon^- = \epsilon^+ = \epsilon$ affect the classifiers' performance on \texttt{adult} dataset. \textsc{SL}: Surrogate Loss. \textsc{GPL}: Group Peer Loss. We highlight substantial improvement of fairness in green and sever violation in red.}
        \label{tab:symmetric_noise_result}
    \scalebox{1}{\begin{tabular}{c  c c c c c}
    \toprule
        Noise $\epsilon$ & Metric & Clean & Corrupt & \textsc{SL} & \textsc{GPL} \\
    \midrule
        \multirow{2}{*}{0.1} & \textit{violation} & $0.47\%$ & $3.91\%$ & \bad{$5.15\%$} & $\good{1.41\%}$ \\
        & \textit{accuracy} & $83.76\%$ & $83.22\%$ & $82.73\%$ & $82.71\%$ \\
    \midrule
        \multirow{2}{*}{0.2} & \textit{violation} & $0.47\%$ & $3.83\%$ & $3.98\%$ & \good{$1.49\%$}\\
        & \textit{accuracy} & $83.75\%$ & $82.08\%$ & $82.54\%$ & $82.16\%$ \\
    \midrule
        \multirow{2}{*}{0.3} & \textit{violation} & $0.47\%$ & \bad{$7.23\%$} & $3.63\%$ & \good{$1.22\%$}\\
         & \textit{accuracy} & $83.76\%$ & $81.36\%$ & $82.01\%$ & $81.24\%$ \\
    \midrule
        \multirow{2}{*}{0.4} & \textit{violation} & $0.47\%$ & \bad{$5.14\%$} & \good{$1.13\%$} & $3.1\%$  \\
        & \textit{accuracy} & $83.76\%$ & $79.58\%$ & $80.62\%$ & $80.21\%$ \\
    \bottomrule
    \end{tabular}}
\end{table}

\subsection{Impact of noise levels on classifier performance.}
We present the results of varying noise rate on the \texttt{adult} data set (with two groups) in Table~\ref{tab:symmetric_noise_result}. We only add symmetric noise to $female$ group and keep the $male$ group clean. ERM is generally robust to symmetric noises when a significant subset of the data is clean (one group in our example), so we do not expect significant accuracy improvement from our methods. We focus on how fairness violation reduces. Observe that, comparing to training with clean data, training on corrupted data substantially increases fairness violations, even for relatively low noise rates. The \textsc{SL} and \textsc{GPL} columns show that our fair ERM approaches can effectively mitigate the biases. This holds true even when increasing the noise rate.

\subsection{Insights on running on data directly, without adding additional noise}\label{app:insight_clean}
We evaluate our algorithm on the clean \texttt{adult} and \texttt{arrest} datasets as shown in Table~\ref{exp:clean_result}. On the \texttt{arrest} dataset, our methods achieve a similar performance of accuracy compared with the Clean baseline, but we do observe a consistent drop of fairness violations on the \texttt{arrest} dataset. The fairness violation of our methods on \texttt{adult} dataset is not as good as that of Clean baseline. This fact may imply the possibility that the \texttt{arrest} dataset contains more human biases in labels than the \texttt{adult} dataset. The small drop in accuracy and (sometimes) in fairness is due to the additional noise estimation step, which introduces another layer of complication - this is the price we pay for dealing with potentially highly noisy labels.
\begin{table}[htb]
    \centering
    \caption{We examine the performance of our methods on the clean \texttt{adult} and \texttt{arrest} datasets. Clean: train a fair classifier directly with equal odds constraint. SL: Surrogate Loss with estimated noise parameters. GPL: Group Peer Loss with estimated noise parameters. The values after $\pm$ are the standard deviation.}
    \label{exp:clean_result}
    \resizebox{\linewidth}{!}{
    \begin{tabular}{c c c c c}
    \toprule
        & \multicolumn{2}{c}{\texttt{adult}} & \multicolumn{2}{c}{\texttt{arrest}} \\
        \cmidrule(r){2-3}\cmidrule(r){4-5}
        Method & \textit{accuracy} & \textit{violation} & \textit{accuracy} & \textit{violation} \\
    \midrule
    Clean &	$83.76 \pm 0.0$	& $0.47 \pm 0.0$ & $65.46 \pm 0.0$ & $4.46 \pm 0.0$ \\
    SL & $76.97 \pm 0.24$ & $3.51 \pm 0.24$ &	$63.07 \pm 0.44$ & $2.90 \pm 0.72$ \\
    GPL	& $81.20 \pm 0.19$ & $3.76 \pm 0.19$ & $64.98 \pm 0.40$ & $1.85 \pm 0.36$ \\
    \bottomrule
    \end{tabular}}
\end{table}
\section{Concluding remarks, limitations and future works}\label{Sec::discussion}
We have demonstrated, both theoretically and empirically, that naively enforcing parity constraints without taking noisy labels into consideration can indeed do harm. Our results show the importance of accounting for group-dependent label-noise when performing ERM subject to fairness constraints. In realistic applications, such as criminal justice and evaluating loan applications, labels are often contaminated by human biases against a certain protected group. The insights gained from this work forewarn decision-makers that improperly mitigating unfairness might do harm on the clean groups. Our two fairness-aware ERM frameworks are an important step toward addressing this problem.

Our work extends a growing body of methods for training classifiers to provide equal opportunity to members of different subgroups within a population. Our new contribution is to address situations where feature and label information for one or more of the subgroups has been recorded less faithfully than for members of other subgroups. Just one example of this, discussed in the text, is the significant disparity in the quality of evaluations for males and females which occur in both medical and academic contexts. These disparities can and do have significant impacts on the quality of life for members of each group, and are well worth addressing.

This work shows how applying existing techniques for mitigating bias in classifiers can actually increase inequality in outcomes, if disparities in the accuracy of training data are not accounted for. We offer new methods for addressing these problems as well. We believe that applying our methods \textit{thoughtfully} will improve existing methods of bias mitigation in machine learning. Our technical solutions and solvers should be of interests to machine learning practitioners/researchers, as well as to policy makers when decided to use classification tools but face a training data with low-quality annotations.

Our work has limitations. Our selection of data sets is limited: we rely on synthetic training data corruption in order to test our methods. This limitation arises from the unavailability of such sensitive data sets for the broader research community. Both this research, and the methods whose shortcomings we have attempted to address, should be re-examined as richer data sets become available for studying disparities in the quality of information recording between members of different subgroups. The lack of relevant data for studying unfairness in machine learning, and the concerns about how to acquire such data while preserving the privacy of people concerned, is itself an important question in this area, although we do not address it in this work.

It is also possible that blind and uncareful application of our approach (by improperly attempting to correct otherwise accurate labels) may in fact create classifiers that produce even greater inequality, or lead to other problems that we have not foreseen. The temptation to apply our methods simply for the purpose of making existing models seem ``more fair,'' especially to unsuspecting downstream users, is very real. We very much discourage the use of our research in this fashion.

Both the limitations and the insights gained through this work underscore an important underlying message: that blind application of bias mitigation techniques in machine learning may do more harm than good.

\begin{acks}
The authors would like to thank Berk Ustun for many inspiring early discussions on the practical scenarios when group-dependent noisy labels can cause harm. 
The authors also thank anonymous reviewers for their constructive comments. This work is partially supported by the National Science Foundation (NSF) under grants IIS-2007951, CCF-2023495, and CCF-1740850.
\end{acks}

\bibliographystyle{ACM-Reference-Format}
\bibliography{reference}

\newpage
\appendix
\onecolumn
\section*{Omitted Proofs}\label{sec:proofs}
\subsection*{Proof of Theorem~\ref{thm:NoiseHarmsAccuracy}}
\begin{proof}
Consider a setting with two identical groups $(X,Y,Z=z)$ and $(X,Y,Z=z')$. Equality of $\TPR$ on the noisy data implies:
\begin{multline}
\Probability(f(X)= +1| Y=+1,Z=z) = \\ \Probability(f(X)= +1 \mid \noisy{Y}= +1, Z=z').\label{equalTPR}
\end{multline}
Since the two groups $(X,Y,Z=z)$ and $(X,Y,Z=z')$ are identical, we have for $s \in \{-1,+1\}$,
\begin{align}
\Probability(f(X)=+1\mid Y=s, Z=z) = \Probability(f(X)=+1\mid Y=s, Z=z').\label{eqn:ident}
\end{align}
Expanding $\Probability(f(X)=+1|Y=+1, Z=z')$ using law of total probability we have
\begin{small}
\begin{align}
    &\Probability(f(X)=+1|Y=+1, Z=z')\nonumber \\
    = & \Probability(f(X)=+1|\mid \noisy{Y}=+1, Y=+1, Z=z') \cdot \Probability(\noisy{Y}=+1 \mid Y=+1, Z=z')\nonumber\\
    &+\Probability(f(X)=+1\mid \noisy{Y}=-1, Y=+1, Z=z') \cdot \Probability(\noisy{Y}=-1 \mid Y=+1, Z=z')\nonumber\\
    = & \Probability (f(X)=+1\mid \noisy{Y}=+1, Z=z') \cdot (1-e)\nonumber \\
     & +  \Probability (f(X)=+1\mid \noisy{Y}=-1, Z=z') \cdot e \label{eqn:positive-2}
\end{align}
\end{small}
Combining Equation (\ref{equalTPR})  with the above,
\begin{small}
\begin{align*}
      &\Probability(f(X)= +1 \mid \noisy{Y}= +1, Z=z')\nonumber\\
     =\ & \Probability(f(X)= +1 \mid Y=+1,Z=z) \tag*{(by Equation~\ref{equalTPR})} \nonumber\\
    =\ & \Probability(f(X)=+1\mid \noisy{Y}=+1, Z=z') \cdot (1-e)
     +  \Probability(f(X)=+1\mid \noisy{Y}=-1, Z=z') \cdot e  \tag*{(by Equation~\ref{eqn:positive-2})} \\
    \Leftrightarrow\ & \Probability(f(X)=+1\mid \noisy{Y}=+1, Z=z') \cdot e \nonumber \\
    =\ & \Probability(f(X)=+1\mid \noisy{Y}=-1, Z=z') \cdot e \\
      \Leftrightarrow\ & \Probability(f(X)=+1\mid \noisy{Y}=+1, Z=z') = \Probability(f(X)=+1\mid \noisy{Y}=-1, Z=z') \numberthis\label{eqn:positive}
\end{align*}
\end{small}
Similarly, we have
\begin{small}
\begin{align}
    & \Probability(f(X)=+1 \mid Y=-1, Z=z')\nonumber \\
    =\ & \Probability(f(X)=+1 \mid \noisy{Y}=+1, Z=z') \cdot e \nonumber \\
    & + \Probability(f(X)=+1 \mid \noisy{Y}=-1, Z=z') \cdot (1-e) \nonumber\\
    =\ &\Probability(f(X)=+1 \mid \noisy{Y}=-1, Z=z')~~ \text{(by Equation \ref{eqn:positive}) }\label{eqn:negative}
\end{align}
\end{small}
Equation (\ref{eqn:positive}) and (\ref{eqn:negative}) jointly imply 
\begin{small}
\begin{align}
& \Probability(f(X)= +1 \mid Y= -1, Z=z)\nonumber \\
=&\Probability(f(X)= +1 \mid Y= -1, Z=z') \tag*{(by Equation \ref{eqn:ident})} \\
=&\Probability(f(X)=+1 \mid \noisy{Y}=-1, Z=z') \tag*{(by Equation \ref{eqn:negative})}\\
=&\Probability(f(X)=+1\mid \noisy{Y}=+1, Z=z') \tag*{(by Equation \ref{eqn:positive})}\\
 =&   \Probability(f(X)= +1 \mid Y= +1, Z=z), \tag*{(by Equation~\ref{equalTPR})} \label{eqn:equal}
\end{align}
\end{small}
thus completing the proof. 
\end{proof}

\subsection*{Proof of Theorem~\ref{thm:effect:noise:e}}
\begin{proof}
Define the following risk measures
\begin{small}
\begin{align*}
    \tilde{R}(f) := \frac{1}{N} \sum_{i=1}^N \tilde{\ell}(f(x_i),\tilde{y}_i)\\
    \hat{R}(f) := \frac{1}{N} \sum_{i=1}^N \hat{\ell}(f(x_i),\tilde{y}_i)
\end{align*}
\end{small}
First, because of Equation \ref{eq:NoiseBounds}, we have
\begin{small}
\begin{align*}
    \tilde{R}(f) &= \frac{1}{N} \sum_{i=1}^N \tilde{\ell}(f(x_i),\tilde{y}_i)\\
    &=  \frac{1}{N} \sum_{i=1}^N  \frac
    {(1-\epsilon^{sgn(-\noisy{y}_i)}_{z_i})\ell(f(\bm{x}_i),\noisy{y}_i)-\epsilon^{sgn(\noisy{y}_i)}_{z_i}\ell(f(\bm{x}_i),-\noisy{y}_i)}
    {1-\epsilon^+_{z_i}-\epsilon^-_{z_i}}\\
        &=  \frac{1}{N} \sum_{i=1}^N  \frac
    {(1-\noisyeps^{sgn(-\noisy{y}_i)}_{z_i})\ell(f(\bm{x}_i),\noisy{y}_i)-\noisyeps^{sgn(\noisy{y}_i)}_{z_i}\ell(f(\bm{x}_i),-\noisy{y}_i)}
    {1-\noisyeps^+_{z_i}-\noisyeps^-_{z_i}}\\
       &+ \frac{1}{N} \sum_{i=1}^N \textstyle{\left(\frac{1-\epsilon^{sgn(-\noisy{y}_i)}_{z_i}}{1-\epsilon^+_z-\epsilon^-_z}- \frac{1-\noisyeps^{sgn(-\noisy{y}_i)}_{z_i}}{1-\noisyeps^+_z-\noisyeps^-_z}\right)}\ell(f(\bm{x}_i),\noisy{y}_i) \\
    &+ \frac{1}{N} \sum_{i=1}^N \textstyle{\left(\frac{\epsilon^{sgn(\noisy{y}_i)}}{1-\epsilon^+_z-\epsilon^-_z}- \frac{\noisyeps^{sgn(\noisy{y}_i)}}{1-\noisyeps^+_z-\noisyeps^-_z}\right)} \ell(f(\bm{x}_i),-\noisy{y}_i)\\
    &=\hat{R}(f) \tag{by definition of $\hat{R}(f)$}\\
    &+ \frac{1}{N} \sum_{i=1}^N \textstyle{\left(\frac{1-\epsilon^{sgn(-\noisy{y}_i)}_{z_i}}{1-\epsilon^+_z-\epsilon^-_z}- \frac{1-\noisyeps^{sgn(-\noisy{y}_i)}_{z_i}}{1-\noisyeps^+_z-\noisyeps^-_z}\right)}\ell(f(\bm{x}_i),\noisy{y}_i) \\
    &+ \frac{1}{N} \sum_{i=1}^N \textstyle{\left(\frac{\epsilon^{sgn(\noisy{y}_i)}}{1-\epsilon^+_z-\epsilon^-_z}- \frac{\noisyeps^{sgn(\noisy{y}_i)}}{1-\noisyeps^+_z-\noisyeps^-_z}\right)} \ell(f(\bm{x}_i),-\noisy{y}_i).
\end{align*}
\end{small}
Using the error bound in Equation (\ref{eq:NoiseBounds}), we have
\begin{small}
\begin{align*}
    &\left|  \frac{1}{N} \sum_{i=1}^N \textstyle{\left(\frac{1-\epsilon^{sgn(-\noisy{y}_i)}_{z_i}}{1-\epsilon^+_z-\epsilon^-_z}- \frac{1-\noisyeps^{sgn(-\noisy{y}_i)}_{z_i}}{1-\noisyeps^+_z-\noisyeps^-_z}\right)}\ell(f(\bm{x}_i),\noisy{y}_i) \right. \\
    & \quad \quad \quad \quad \left. + \frac{1}{N} \sum_{i=1}^N \textstyle{\left(\frac{\epsilon^{sgn(\noisy{y}_i)}}{1-\epsilon^+_z-\epsilon^-_z}- \frac{\noisyeps^{sgn(\noisy{y}_i)}}{1-\noisyeps^+_z-\noisyeps^-_z}\right)} \ell(f(\bm{x}_i),-\noisy{y}_i) \right| \\
    &\leq \frac{1}{N} \sum_{i=1}^N  \left| \textstyle{\left(\frac{1-\epsilon^{sgn(-\noisy{y}_i)}_{z_i}}{1-\epsilon^+_z-\epsilon^-_z}- \frac{1-\noisyeps^{sgn(-\noisy{y}_i)}_{z_i}}{1-\noisyeps^+_z-\noisyeps^-_z}\right)}\right| \ell(f(\bm{x}_i),\noisy{y}_i) \\
    & \quad \quad \quad \quad + \frac{1}{N} \sum_{i=1}^N \left|  \textstyle{\left(\frac{\epsilon^{sgn(\noisy{y}_i)}}{1-\epsilon^+_z-\epsilon^-_z}- \frac{\noisyeps^{sgn(\noisy{y}_i)}}{1-\noisyeps^+_z-\noisyeps^-_z}\right)}\right| \ell(f(\bm{x}_i),-\noisy{y}_i) \\
    &\leq \frac{1}{N} \sum_{i=1}^N \tau \bar{\ell} + \frac{1}{N} \sum_{i=1}^N \tau \bar{\ell} \\
    &= 2\tau \bar{\ell}
\end{align*}
\end{small}
Then we conclude that $\forall f$
\begin{align}
    | \tilde{R}(f) - \hat{R}(f)| \leq 2\tau \bar{\ell} \label{eqn:bound}
\end{align}
This enables us to obtain the following bound
\begin{small}
\begin{align*}
    \tilde{R}(\hat{f}^*) - \tilde{R}(\tilde{f}^*) &
    \leq \hat{R}(\hat{f}^*) + 2 \tau \bar{\ell} - \tilde{R}(\tilde{f}^*) \tag*{(by Equation \ref{eqn:bound})}\\
    & \leq \hat{R}(\tilde{f}^*)- \tilde{R}(\tilde{f}^*)+ 2 \tau \bar{\ell} \tag*{(because of the optimality of $\hat{f}^*$ with respect to $\hat{R}(f)$)}\\
    & \leq 2 \tau \bar{\ell}+ 2 \tau \bar{\ell} && \tag*{(by Equation \ref{eqn:bound})}\\
    & = 4\tau \bar{\ell}.
\end{align*}
\end{small}
\end{proof}

\end{document}